\documentclass[runningheads]{llncs}
\usepackage{graphicx}
\usepackage{relsize}
\usepackage{caption}
\usepackage{subcaption}

\begin{document}
\title{Adaptive Explainable Continual Learning Framework for Regression Problems with Focus on Power Forecasts}

\titlerunning{Explainable Continual Learning Framework for Regression Problems}
%
\author{Yujiang He
}
\authorrunning{Y. He}
%
\institute{Intelligent Embedded Systems Lab, University of Kassel, Germany
\email{yujiang.he@uni-kassel.de}\\
\url{https://www.uni-kassel.de/eecs/ies}}
\maketitle              
\begin{abstract}
Compared with traditional deep learning techniques, continual learning enables deep neural networks to learn continually and adaptively.
Deep neural networks have to learn unseen tasks and overcome forgetting the knowledge obtained from previously learned tasks as the amount of data keeps increasing in applications.
This article proposes two continual learning application scenarios, i.e., the target-domain incremental scenario and the data-domain incremental scenario, to describe the potential challenges in this context.
Based on our previous work regarding the CLeaR framework, which is short for continual learning for regression tasks, models will be enabled to extend themselves and to learn data successively.
Research topics are related, but not limited, to developing continual deep learning algorithms, strategies for non-stationarity detection in data streams, explainable and visualizable artificial intelligence, etc.
Moreover, the framework- and algorithm-related hyperparameters should be dynamically updated in applications.
Forecasting experiments will be conducted based on power generation and consumption data collected from real-world applications.
A series of comprehensive evaluation metrics and visualization tools are applied to access the experimental results.
The proposed framework is expected to be generally applied to other constantly changing scenarios. 
\keywords{Continual Learning \and Non-stationary Data Stream \and Explainable AI \and Power Forecasts \and Smart Grids }
\end{abstract}
\section{Introduction}
\label{sec:introduction}
Organic Computing (OC) aims at constructing a trustworthy and safe system with the properties, such as self-organization, self-adaption, and self-protection~\cite{tomforde2017organic,muller2017organic}.
The system can interact with humans in an organic manner, i.e., collecting and analyzing data from the sensors surrounding humans, and adjusting itself to fulfill the instant requirements of users.
To achieve the goal of Organic Computing, we should consider the flexible deployment and self-evolution of the system when developing it.
It rises a challenge in common deep learning technology.
Therefore, Continual Learning (CL) has been proposed as an initiative in the research field of deep learning for overcoming the challenge.

Training successful deep neural networks usually depend on a massive amount of data. 
With one-time design and training, deep neural network models can be easily deployed to solve specific problems.
In recent years, a lot of research has proven that such a traditional data-driven training method can quickly optimize hyperparameters of deep neural networks with the help of massive datasets and supercomputing resources.
Models can obtain or even exceed human-level cognitive skills in numerous application scenarios.
However, even such a common training method has three main disadvantages:
\begin{enumerate}
    \item \textbf{Insufficient samples}: a training dataset with sufficient meaningful samples is the prerequisite for training a successful model.
    Data collection and preprocessing are extremely time-and-money-consuming.
    Meanwhile, the tedious process can prolong the preparation phase in a practical project.
    Sometimes, we have to postpone the beginning of training models due to a lack of data.
    \item \textbf{Static models}: in this training setting, it is always assumed that the underlying data generation process is static.
    Therefore, we can evaluate models' generalization by comparing the errors among training, validation, and test datasets in the training phase.
    However, this static model is not always applicable to the constantly evolving and developing world.
    In this article, the context can be defined as a non-stationary scenario, where the generative probabilistic distribution of both the input data $P(X)$ or the target data $P(Y|X)$ changes over time. 
    Changes can be grouped into three families, i.e., short periodical, long periodical, or non-periodical~\cite{he2021application}.
    Insufficient samples involving periodical changes can restrict the model from obtaining the information of the entire sample space during training.
    Non-periodical changes could be caused by changes in the objective environment, broken physical devices, or unavoidable measurement errors.
    Deep neural network models should continually learn the data with these periodical changes to improve their cognition, similar to the learning process of humans.
    Because the non-periodical changes are hardly predictable and repetitive, we need to detect them and make the correct decisions for processing them.
    \item \textbf{Post-deployment changes}: the structure of the deep neural network model is generally fixed after deployment.
    This setting is unrealistic and inflexible in real-world applications because new targets can appear as the application environment keeps evolving.
    The model should extend its structure by increasing the number of outputs in this case.
    A new target can be a new label in classification tasks or a new predicted object in regression tasks.
    To take power forecasts in the context of smart power girds as an example, we can train models to provide forecasts regarding energy supply and demand for managing a regional power grid.
    With the extension of the power grid, new power generators and consumers must be added to the list of forecasting targets.
    Training an individual model for the new target might also be a solution.
    However, in this case, we have to reconsider the first problem, i.e., we can not start training in a traditional way until sufficient samples are collected.
\end{enumerate}
\textbf{Continual Learning}, also known as Continuous Learning, Incremental Learning, Sequential Learning, or Lifelong Learning,  aims to solve multiple, related tasks and lead to a long-term version of machine learning models.
While the term is not well consolidated, the idea is to enable models to continually and adaptively learn about the world and overcome catastrophic forgetting.
The knowledge of models can be developed incrementally and become more complex.
Catastrophic forgetting refers to the phenomenon that models could forget the knowledge for solving old tasks while learning new tasks~\cite{mccloskey1989catastrophic,ratcliff1990connectionist}.
This forgetting problem raises a more general problem in traditional neural networks, i.e., the so-called stability-plasticity dilemma~\cite{mermillod2013stability}, which means that models should find a trade-off between accumulation and integration of new knowledge and retention of old knowledge.
Numerous valuable research work focused on CL algorithms~\cite{chaudhry2018riemannian,li2017learning,van2018generative,zenke2017continual}, application scenarios~\cite{lomonaco2017core50}, evaluation metrics~\cite{diaz2018don} for classification tasks, etc.
However, the necessity of CL for regression tasks seems to be ignored.

This article can be viewed as an abstract for the author's Ph.D. thesis with a focus of continual learning applied to the regression scenario, especially in the field of power forecasts. 
Meanwhile, the proposed continual learning framework is also expected to be a promising solution of self-evolution of systems to contribute to OC.
The main contributions of the final thesis can be summarized as follows:
\begin{enumerate}
    \item Prove the necessity and importance of CL for regression tasks.
    \item Give an overview of the relevant research literature, including, but not limited to, CL algorithms, detection of novelty and non-stationarity in the data streams and explainable artificial intelligence (AI);
    \item Explore the applicability of well-known CL algorithms for regression problems.
    \item Analyze the shortcomings of common experimental setups as well as restrictions of general evaluation metrics.
    \item Summarize relevant research challenges being faced for our proposed CL framework~\cite{he2021clear} and develop it further.
    \item Develop explainable CL using visualization techniques and comprehensive evaluation metrics.
    \item Show how the CL framework works in power forecasting experiments with real-world datasets.
\end{enumerate}
The remainder of this article will start with an overview of the requirements and relevant research questions of CL for regression problems in Section~\ref{sec:cl_for_regression}.
In Section~\ref{sec:clear_framework}, we will propose a visualizable CL framework for regression problems and introduce the application in the two proposed CL scenarios with instances.
Then we will present three experimental datasets, which can be used to design power forecasts experiments to assess the proposed solutions.
This article will end with a brief conclusion.
\section{Continual Learning for Regression}
\label{sec:cl_for_regression}
\subsection{Continual Learning Scenarios}
\label{subsec:scenarios}
CL has been widely applied to classification problems for learning new tasks sequentially and retaining the obtained knowledge.
Three CL scenarios for classification problems are proposed in~\cite{lomonaco2017core50}, focusing on object recognition:
\begin{itemize}
    \item \textbf{New Instances}: new samples of the previously known classes are available in the subsequent batches with novel information, such as new poses or conditions.
    In other words, these new samples own novel information but still belong to the same labels.
    Models need to keep extending and accumulating knowledge regarding the learned labels.
    \item \textbf{New Classes}: new samples belong to unknown classes.
    In this case, the model should be able to identify the objects of new classes as retaining the accuracy of old ones.
    \item \textbf{New Instances and Classes}: new samples belong to both known and new classes.
\end{itemize}
Therefore, a new task in the context of classification problems can be defined as learning new instances belonging to known labels or learning to recognize new labels.
However, one obvious difference between classification and regression problems is the models' targets, which are discrete labels in classification and continuous values in regression.

Two CL scenarios for regression are proposed in~\cite{he2020continuous}:
\begin{itemize}
    \item \textbf{Data-domain incremental (DDI) scenario} refers to the situation where the underlying data generation process is changing over time due to the non-stationarity of the data stream.
    Either the change of the probability distribution regarding the input data $P(X)$ or the target $P(Y|X)$ can trigger updating the model trained on data from the outdated generation process.
    The model learns to extract latent representations of the input data from a changed generation process in the updating phase.
    Moreover, the model needs to adjust its weights to find a new proper mapping between the new latent representations to the targets.
    The changes of distribution could result from insufficient samples in the pre-training process or external objective factors.
    
    \item \textbf{Target-domain incremental (TDI) scenario} refers to the situation where the structure of the network model is extended as the number of prediction targets increases.
    Assume that using a multi-output deep neural network to forecast several independent targets based on the same input data, the neural network owns a shared hidden sub-network for learning non-linear latent representations and multiple-output sub-networks for prediction.
    The model will add a new sub-network for the new target when it appears.
    The TDI scenario is a joint research topic among multi-task learning, transfer learning and continual learning.
    On the one hand, the obtained knowledge of the shared network can be transferred to train the additional sub-network quickly, even without sufficient samples.
    On the other hand, CL algorithms can avoid decreasing the prediction accuracy of the previously handled tasks while learning the new task by utilizing the free weights of the shared network.
    The free weights are generally unimportant for other targets.
\end{itemize}

\begin{figure}[ht]
 \centering
 \includegraphics[width=0.98\textwidth]{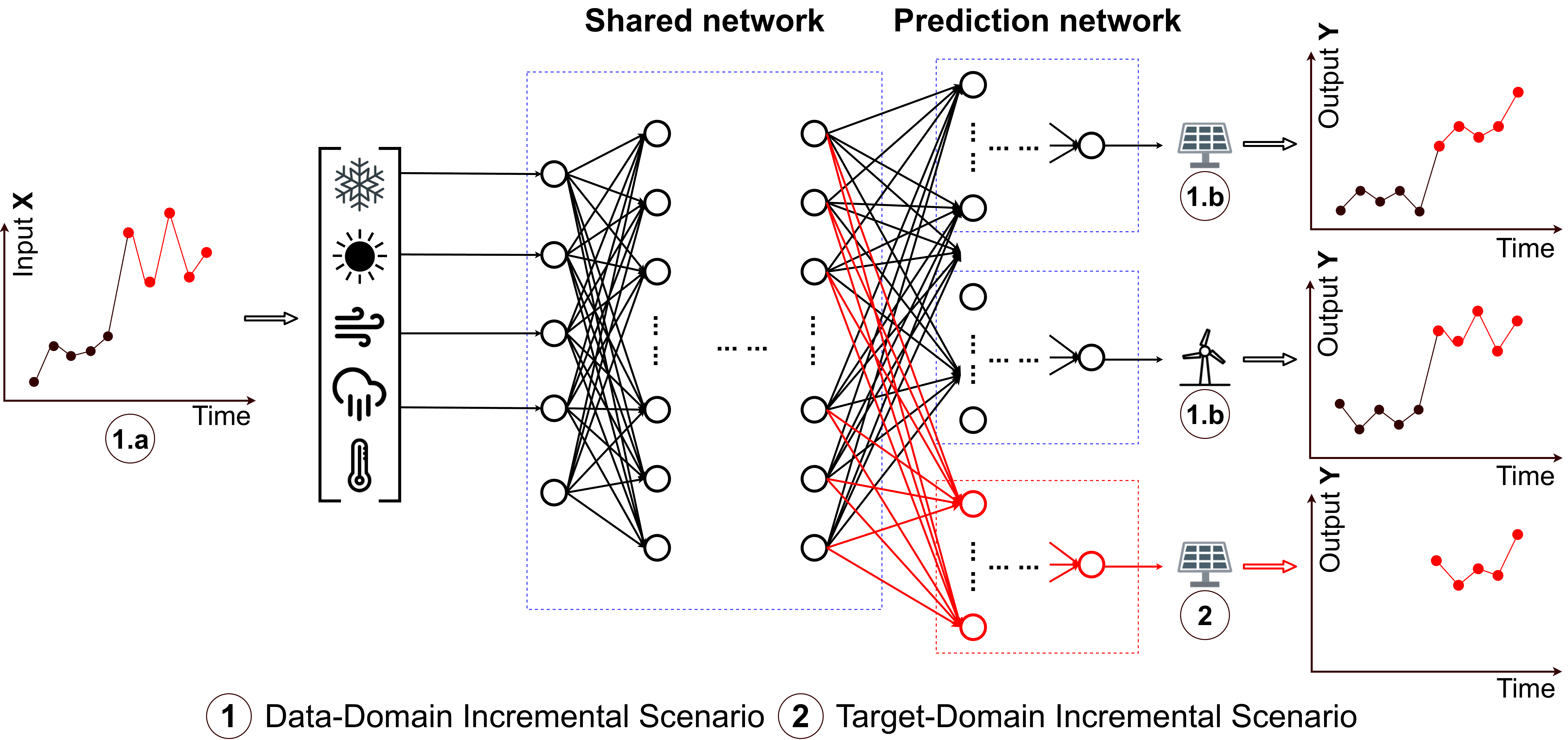}
 \caption{An illustration of a neural network applied to learn new tasks in both CL application scenarios. Red dots and lines indicate the new samples and new sub-networks. The cases, marked as \textbf{1.a} and \textbf{1.b}, correspond to the DDI scenario, where the distribution $P(X)$ and the mapping $P(Y|X)$ change respectively. The right case, marked as \textbf{2}, corresponds to the TDI scenario where new targets are added to the prediction list.} 
 \label{fig:2scenarios}
\end{figure}

For example, renewable power generation can be predicted based on regional weather conditions.
Figure~\ref{fig:2scenarios} illustrates the two proposed CL scenarios to power forecasts of regional renewable energy generators.
As described above, the model has a shared network to learn the latent representations and several prediction sub-networks for these targets.

The weather conditions are time-variant features, which fluctuate periodically over time.
Therefore, a gradual change can exist in the weather data $X$ due to climate change, dynamic behavior of noise from the weather prediction model, or other foreseeable factors.
Such smooth change is usually referred to as concept drift~\cite{gruhl2021novelty}.
The case in Fig.~\ref{fig:2scenarios}, marked as \textbf{1.a}, visualizes this challenge, which will lead to a negative impact on prediction accuracy, especially when sufficient samples are unavailable for pre-training.

The cases of \textbf{1.b} are corresponding to power generation capability, which is time-dependent and can be affected by, such as, upgrading or aging of the device in the long term or the changes in the environment.
Besides, residential power demand forecasting is another example that needs to be considered in this scenario.
Generally, we predict the overall power demand of a residential area in a low-voltage power grid rather than the power demand of every single consumer in this region.
The mapping $P(Y|X)$ is sensitive to the change of these consumers' power demand or consumption habits.
Sometimes, we have to update the prediction model due to these unpredictable factors.
In the DDI scenario, models should continually collect data and accumulate knowledge by learning newly collected data.
Regarding the TDI scenario in Fig.~\ref{fig:2scenarios}, the prediction sub-network for an additional photovoltaic generator is added to the prediction model.

In the proposed setting, a new task can thus be defined in one of two ways:
\begin{enumerate}
    \item Learning non-stationarity of the data stream, including the input data generation and the output data generation with given inputs.
    \item Integrating new sub-networks to the existing model for predicting new targets without any negative effects on the prediction accuracy of other known targets. 
\end{enumerate}
The red dots and lines in Fig.~\ref{fig:2scenarios} correspond to the new tasks in both scenarios, respectively.

\subsection{Research Questions}
\label{subsec:questions}
In the common CL experimental setting for classification tasks, the dataset $D$ contains $T$ disjoint subsets, each of which is assumed to be separately sampled from an independent and identical distribution.
One subset represents a task, and the dataset $D$ is not independently and identically distributed, which is different from traditional supervised learning.
Neural network models need to learn these unseen, independent tasks sequentially as the identification information of these tasks is given.
Some CL algorithms allow models to revisit the previously learned tasks without restriction while learning a new one.
It is called replay CL strategy which will be introduced below in detail.
The replayed data can be a subset of samples storing the previous tasks in raw format or generated by a generative model.
This setting can not represent the case in the real world, though it is feasible to evaluate and compare various CL algorithms.

First, the appearance of a new task is usually unpredictable in real-world applications, which means that the prior knowledge of the new task is unavailable.
A detection mechanism should be set to identify the appearance and the type of the new tasks.
Second, data streams in real applications are infinite and contain both known and unknown tasks which might not appear separately and orderly.
The model could identify the new task and update itself immediately or store the samples until there is a sufficient amount of samples for an update, which depends on the adopted updating strategy.
Third, unrestricted retraining on old tasks enables the model to remember the obtained knowledge and might be prone to overfitting.
Besides, storing all old tasks might be a burden to storage overhead and violate privacy laws.

Farquhar et al. introduce five core desiderata for evaluating CL algorithms and designing classification experiments~\cite{farquhar2018towards}.
It inspires us to propose five suggestions for designing CL regression experiments in ~\cite{he2021clear}:
\begin{enumerate}
    \item New tasks resemble the previous tasks;
    \item The neural network model utilizes the single output for predicting the corresponding target and learning the changes in DDI scenario;
    \item New tasks appear unpredictably in DDI scenario, and the prior knowledge regarding the appearance should be identified rather than informed;
    \item Considering privacy law and revisiting the previous dataset with restriction;
    \item Learn more than two tasks continually, either in DDI scenario or TDI scenario. 
\end{enumerate}
These suggestions will guide the development of updating methods and the design of experiments.
Furthermore, the following research questions should be answered in the proposed thesis.

\subsubsection{Question 1: When to trigger an update?}
The trigger condition is the prerequisite for CL and determines the starting point of an update.
According to the definition of a new task in Section~\ref{subsec:scenarios}, this question is more valuable to research in the DDI scenario.
New tasks of the DDI scenario appear unpredictably, compared to the TDI scenario where new tasks depend on the objective requirements of projects and are added manually.
For answering this question, my research will focus on novelty detection (concept shift and drift) using deterministic and probabilistic methods.
For example, in~\cite{he2021clear}, the trigger condition depends on the number of newly collected novel samples.
Besides, updating could also be triggered due to the estimation of new samples' entropy.
The design of update trigger conditions is the first significant step affecting the updating results and the model's future performance.
\subsubsection{Question 2: How to update models?}
Updating methods are the core for learning tasks sequentially and continually.
CL algorithms can be briefly categorized into three groups~\cite{delange2021continual} depending on how data is stored and reused:
\begin{itemize}
    \item \textbf{Regularization-based approaches}: The goal of these algorithms is to reduce storage demand and to prioritize privacy by avoiding to revisit old data as the neural network learns new tasks. 
    A penalty regularization term is added to the loss function for consolidating the weights which are important for previous tasks.
    Delange et al. further divided these approaches into data-focused approaches~\cite{li2017learning,zhang2020class,rannen2017encoder} and prior-focused approaches~\cite{kirkpatrick2017overcoming,zenke2017continual,chaudhry2018riemannian}.
    
    \item \textbf{Replay approaches}: These approaches prevent forgetting through replaying the previous samples stored in raw format or generated by a generative model.
    Both can be named as rehearsal approaches~\cite{rebuffi2017icarl,de2020continual} and pseudo rehearsal approaches~\cite{shin2017continual,lavda2018continual}, respectively. 
    The model's inputs contain the subset of these previous samples and the unseen samples.
    The former is viewed as a constrain during optimizing the loss function and the latter is the new task for continual learning.
    
    \item \textbf{Parameter isolation approaches}: The concept of these approaches is to arrange a subset of the model's parameters to a new task specifically.
    For example, one can adjust the model's structure by expanding a new branch to learn a new task if no constraint is required for the model's size~\cite{rusu2016progressive}. 
    Alternatively, the shared part of the network for all tasks can stay static.
    The parts regarding the previous tasks are marked out while the network learns new tasks~\cite{mallya2018packnet,fernando2017pathnet}.
\end{itemize}
Some previous works~\cite{Kurle2020Continual,minka2009virtual,nguyen2018variational} solve the forgetting problem using Bayesian Neural Networks, which can also be grouped into one of the above families.

Note that not all well-known CL can directly be applied to regression tasks.
For example, Li et al. use Knowledge Distillation loss~\cite{hinton2015distilling} in their Learning Without Forgetting (LWF) algorithm to consolidate the obtained knowledge for previous tasks~\cite{li2017learning}. 
The loss function is a variant of cross-entropy loss, which is inapplicable for regression tasks.
Thus, the author plan to review these well-known algorithms and then analyze their advantages and applicability.
Moreover, further work proposes novel CL algorithms based on the current CL and the proposed experimental setup.
Another interesting topic is ensemble CL, which investigates the collaboration of various CL algorithms to improve the models' performance.

\subsubsection{Question 3: How to evaluate the updated models?}
Common evaluation metrics for regression tasks, such as Mean Square Error (MSE), Root Mean Square Error (RMSE), and Mean Absolute Error (MAE), can assess the fitting and prediction ability of the model.
However, more specific metrics are required for evaluating the updated models comprehensively in the CL setting.
In~\cite{he2021application,he2021clear}, the models are evaluated in terms of fitting error, prediction error, and forgetting ratio.
We also consider training time as a significant evaluation factor~\cite{he2020continuous}, especially in real-time applications.
Besides, D{\'\i}az-Rodr{\'\i}guez et al.~\cite{diaz2018don} propose algorithm ranking metrics according to different desiderata, including accuracy, forward/backward transfer, model size efficiency, sample storage size efficiency, and computational efficiency.
Moreover, they fuse these desiderata to a single CL score for ranking purposes.
A series of wide-ranging evaluation metrics can make CL explainable.
It is the basis of visualizing the updating process and dynamically adjusting hyperparameters.

\subsubsection{Question 4: How to explain the updating process?}
The training process of either typical supervised learning or CL is generally a black box, where results of the solution are untransparent and incomprehensible for humans.
Explainable artificial intelligence, also called XAI in literature, refers to the technique used to help humans understand and trust machine learning models' results.
It has been applied to many sectors, such as medical diagnoses~\cite{holzinger2017we} and text analysis~\cite{qureshi2019eve}.

Due to stochasticity in re-learning, some updates could fail and lead to a worse predictive ability. 
XAI can visualize the updating process and interpret the reasons for failures.
Experts can monitor the updating process and analyze the updated model based on the given evaluation criterion.
Furthermore, they can more easily decide to accept successful updates or reject failures, and take the following actions.
For example, rolling back the failed model to the previous version, assembling multiple updated models for ensemble learning, or adjusting hyperparameters dynamically for a further update.
\section{Visualizable Continual Learning Framework for Regression Tasks}
\label{sec:clear_framework}
\begin{figure}[!t]
 \centering
 \includegraphics[width=0.98\textwidth]{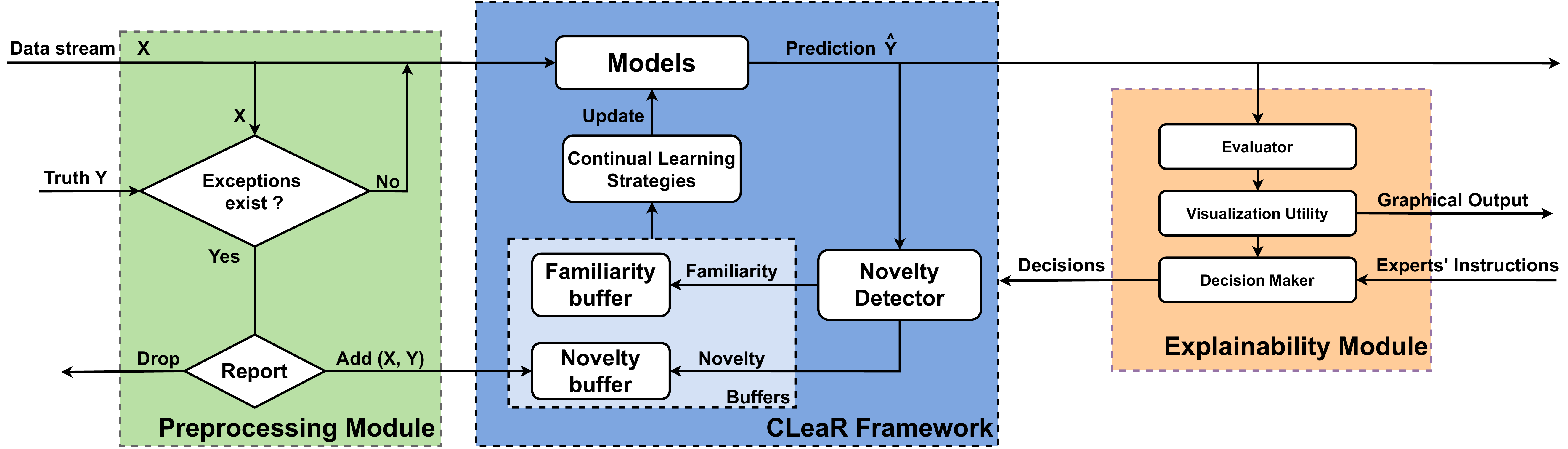}
 \caption{An illustration of the visualizable continual learning frameowrk for regression tasks. It consists of three parts: (1) preprocessing module, (2) CLeaR framework, and (3) explainability module. The dashed line means that the \textbf{Truth Y} is optional.} 
 \label{fig:clear_workflow}
\end{figure}

The V-CLeaR framework (\textbf{V}isualizable \textbf{C}ontinual \textbf{Lea}rning for \textbf{R}egression tasks) in Fig.~\ref{fig:clear_workflow} consists of three main parts: (1) preprocessing block, (2) CLeaR framework, and (3) explainability utility.
The preprocessing block is responsible for processing the incoming data, including removing outliers, filling missing values, and scaling.
Due to concept drift in the data stream, the parameters of the used scaler might have to be updated, for example, the maximum and minimum of the min-max scaler or the mean and variance of the standard scaler.

CLeaR is a continual learning framework based on buffered data~\cite{he2021clear} that is categorized as novelty and familiarity by the deterministic or probabilistic novelty detector and stored in the corresponding buffer.
The novelty from the infinite data stream indicates what the trained model cannot predict accurately and should continually learn.
The familiarity is defined as data that the model is familiar with.
It could be obtained from the infinite data stream or the stored historical samples in a raw format, or generated by a generative model.
Storage and usage of these buffered data are dependent on the adopted CL strategies.
\begin{figure}[!t]
\centering
    \begin{subfigure}[t]{0.98\textwidth}
        \centering
        \includegraphics[width=0.98\textwidth]{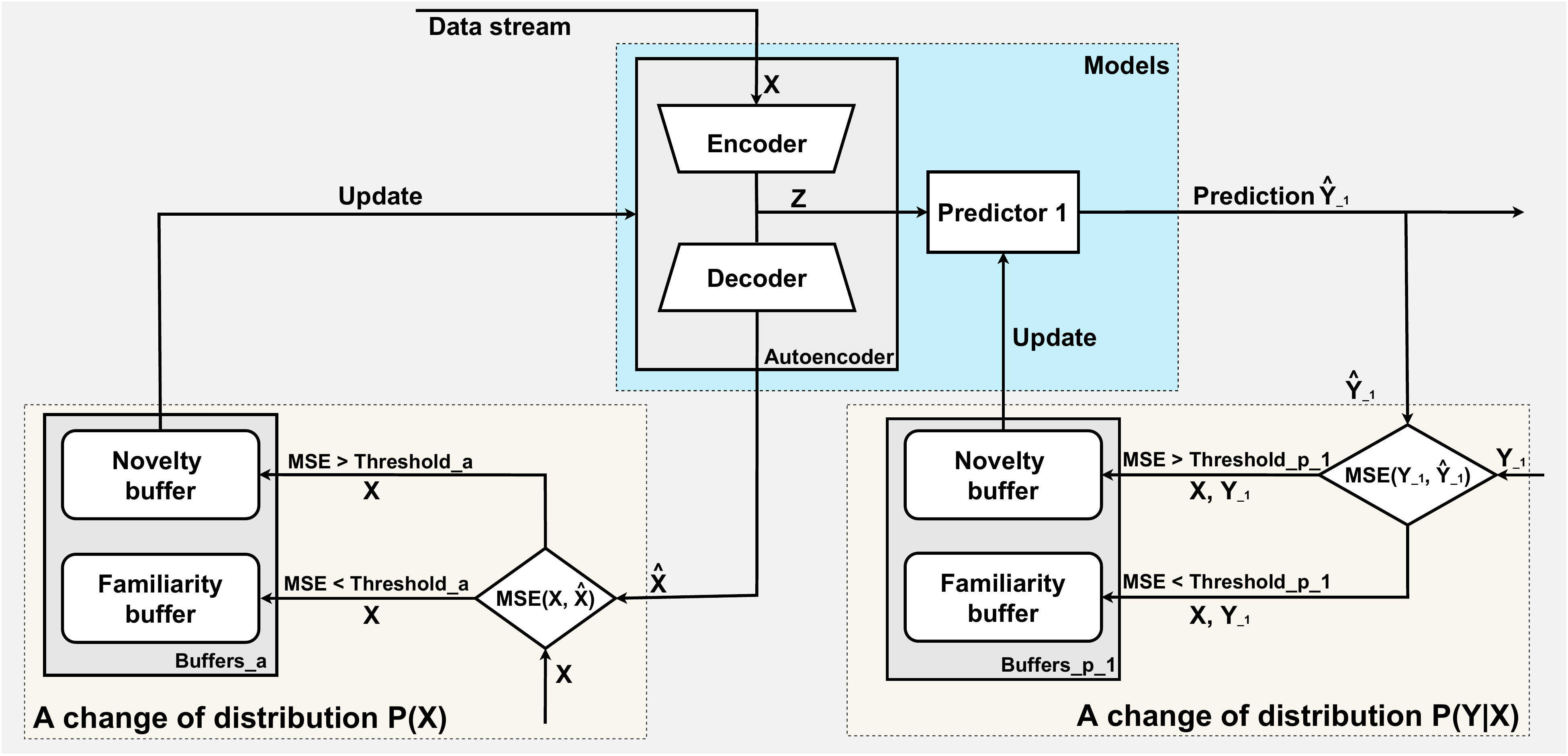}
        \caption{The CLeaR instance in the \textbf{DDI scenario}.}
        \label{fig:clear_ddi}
    \end{subfigure}
    \hfill
    \begin{subfigure}[t]{0.98\textwidth}
        \centering
        \includegraphics[width=0.98\textwidth]{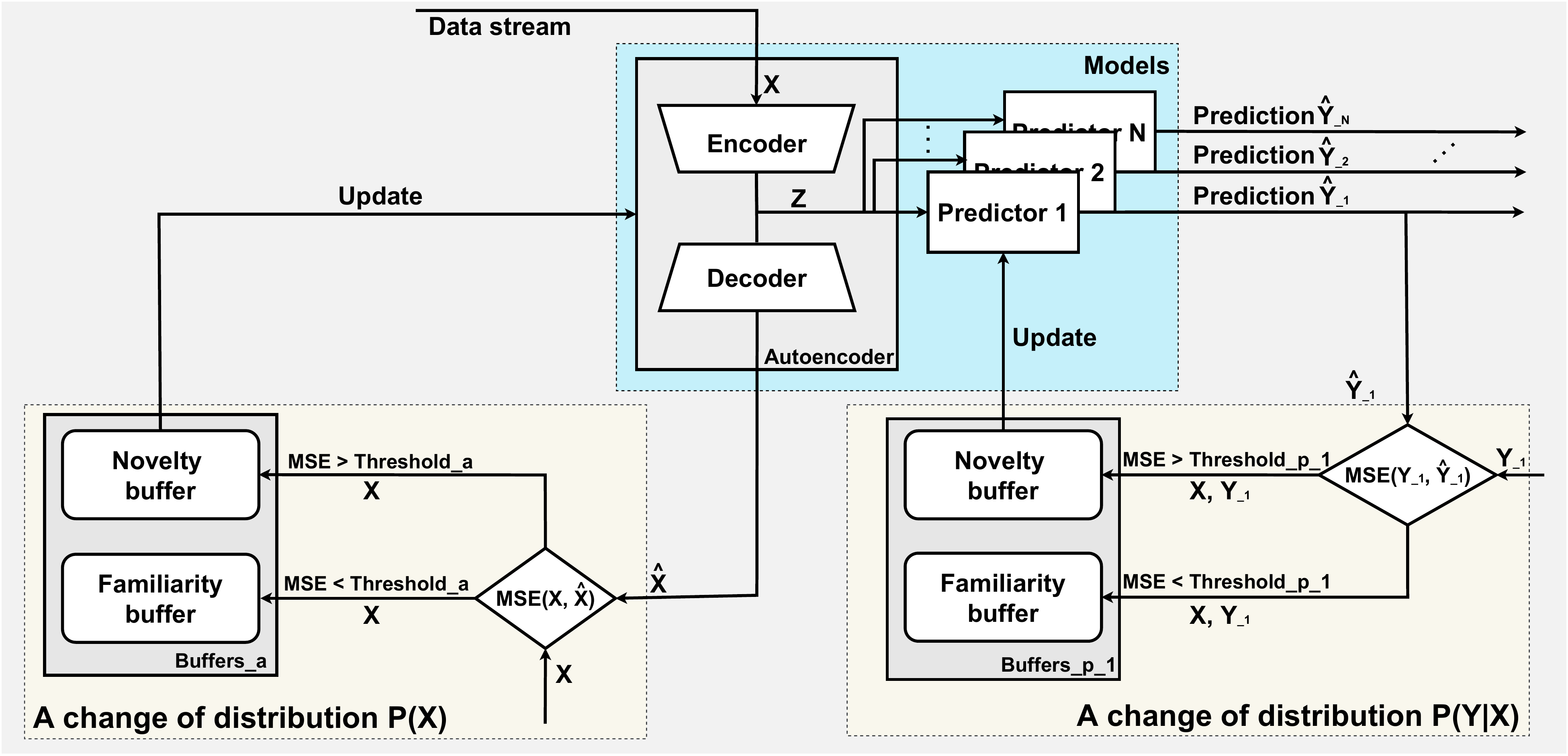}
        \caption{The CLeaR instance in the \textbf{TDI scenario}.}
        \label{fig:clear_tdi}
    \end{subfigure}
    \caption{The illustrations of the CLeaR instances in the two CL scenarios, originally derived from~\cite{he2021application}. \textbf{Threshold\_a}, \textbf{Buffer\_a}, \textbf{Threshold\_p\_1} and \textbf{Buffer\_p\_1} are the framework-related hyperparameters for the autoencoder and the predictor 1, respectively. All sub-networks that need to be updated in application own a series of framework-related components and hyperparameters.}
    \label{fig:clear_scenario}
\end{figure}

Figure~\ref{fig:clear_scenario} illustrates the CLeaR instances in both proposed CL scenarios.
In this instance, the model consists of an autoencoder as the shared network for extracting the latent representations of the input and fully connected networks as the predictors.
Meanwhile, the two sub-networks are used for detecting the changes in the input distribution $\mathrm{P(X)}$ and the goal distribution $\mathrm{P(Y|X)}$, respectively.
Either novelty or familiarity is determined by comparing the MSE between the prediction $\mathrm{\hat{Y}}$ (or reconstruction $\mathrm{\hat{X}}$) and the ground-truth $\mathrm{Y}$ (or $\mathrm{X}$) to the preset dynamically changeable threshold.
The novelty buffer has a limited size, and the familiarity buffer is infinite.
Updating the sub-network is triggered when the corresponding novelty buffer is filled.
After updating the sub-network, the corresponding threshold will be adjusted depending on the updating results, and the CLeaR framework's both buffers will be emptied.
The core of this framework is the flexibility and customizability of these modules, including the novelty detector, storage of the data, the available CL strategies, and the type of neural network models.
Users can select the optimal components of the framework for their own applications.

The explainability utility is designed as a visualization tool focusing on visualizing the updating process and explaining the updated model using the proposed evaluation metrics.
The updating process is also supervised by experts, who can input instructions to assist the model in making decisions for the next move.
Here a decision is defined as what will affect the CLeaR instance to take the following actions.
For example, the CL-algorithm-related hyperparameters are adjusted for re-updating when the current updating results are not ideal.
The framework-related hyperparameters will be changed to make a trade-off between forgetting and prediction in the future update.
Considering these factors, such as storage and computational overhead, experts can decide to store or drop the updated models.
The development of the V-CLeaR framework can answer the four research questions listed in Section~\ref{subsec:questions}.

\section{Datasets \& Experiments}
\label{sec:data_experiments}
In this thesis, the author plan three experiments in the context of power forecasts based on three real-world public datasets to assess the framework's performance.
The datasets contain time series data, the probabilistic distributions of which change periodically/non-periodically over time.
Besides, the mapping relationship between the weather conditions (inputs) and the power measurements (outputs) might also change in a long-term usage due to aging/damaging/upgrading of devices.
Therefore, the datasets are more useful for analyzing and evaluating the performance of the proposed continual learning framework and can highlight its advantages.

In the remainder of this section, the author will briefly introduce the selected datasets and the experimental setup.
\subsection{Wind Power Generation Forecasts}
\label{subsec:wind}
\begin{figure}[!t]
 \centering
 \includegraphics[width=0.6\textwidth]{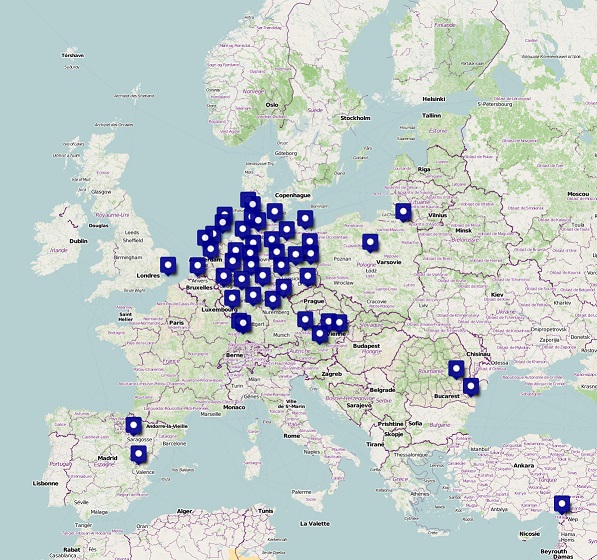}
 \caption{The locations of the European wind farms dataset.} 
 \label{fig:windpark}
\end{figure}

The EuropeWindFarm dataset~\cite{gensler2016} contains the day-ahead power generation of 45 wind farms (off- and onshore) scattered over the European continent, as shown in Fig.~\ref{fig:windpark}.

The dataset contains hourly averaged wind power generation time series for two consecutive years and the corresponding day-ahead meteorological forecasts provided by the European Centre for Medium-Range Weather Forecasts (ECMWF) weather model.
The meteorological features contain (1) wind speed in 100m height, (2) wind speed in 10m height, (3) wind direction (zonal) in 100m height, (4) wind direction (meridional) in 100m height, (5) air pressure, (6) air temperature, and (7) humidity.
All features are scaled between 0 and 1.
Additionally, the power generation time series is normalized with the wind farm's respective nominal capacity to enable a scale-free comparison and to mask the original characteristics of the wind farm.
The dataset is pre-filtered to discard any period of time longer than 24 hours in which no energy has been produced, as this is an indicator of a wind farm malfunction.

V-CLeaR models can be built with the weather features as the input to predict the wind power generation at the corresponding time points~\cite{he2021clear,gensler2016deep}.
One model is trained for each prediction target, i.e., the wind power generator.
This experiment can simulate the DDI scenario.

\subsection{Solar Power Generation Forecasts}
\label{subsec:solar}
\begin{figure}[t]
 \centering
 \includegraphics[width=0.6\textwidth]{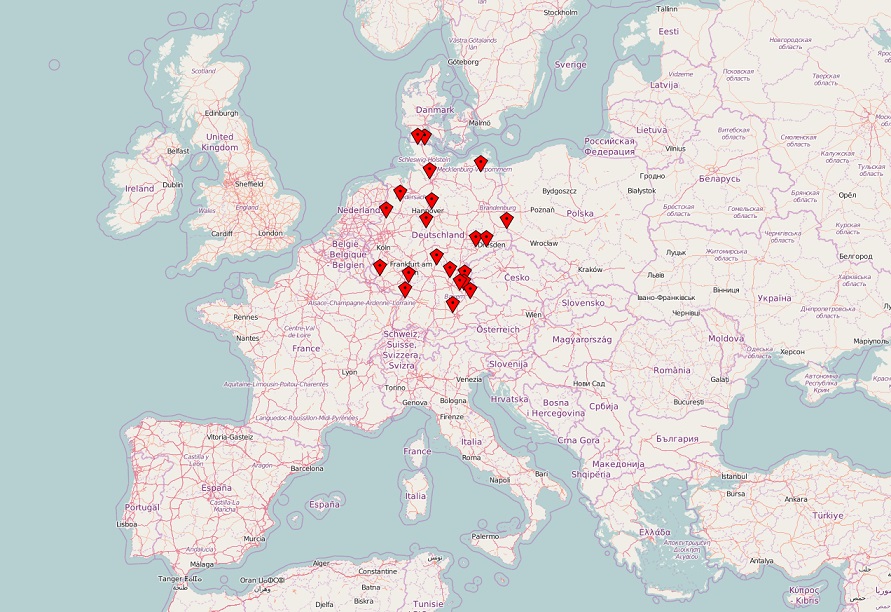}
 \caption{The locations of the German solar farms dataset.} 
 \label{fig:solarfarm}
\end{figure}

The GermanSolarFarm dataset~\cite{solarfarm2016,gensler2016deep} contains 21 photovoltaic (PV) facilities in Germany, as shown in Fig.~\ref{fig:solarfarm}.

Their installed nominal power ranges between 100kW and 8500kW. 
The PV facilities range from PV panels installed on rooftops to fully-fledged solar farms. 
Historical numeric weather prediction (NWP) data and the produced power in a three-hour resolution for 990 days are available for each facility.
The weather prediction series in the dataset are scaled between 0 and 1 using the min-max normalization. 
Besides, there are three temporal features, the hour of the day, the month of the year, and the season of the year, which are normalized in the range of 0 and 1 using cosine and sine coding.
The target variable, i.e., the measured power generation, is normalized using the nominal output capacity of the corresponding PV facility.
Therefore, it allows the comparison of the forecasting performance without taking the size of the PV facilities into account.

The experimental setup is the same as the setup of the wind power generation forecasts experiment. Namely, the NWP features are used as the input of V-CLeaR instances for forecasting each generator in the DDI scenario.

\subsection{Power Supply and Demand Forecasts in a Regional Power Grid}
\label{subsec:power_grid}
\begin{figure}[t]
 \centering
 \includegraphics[width=0.6\textwidth]{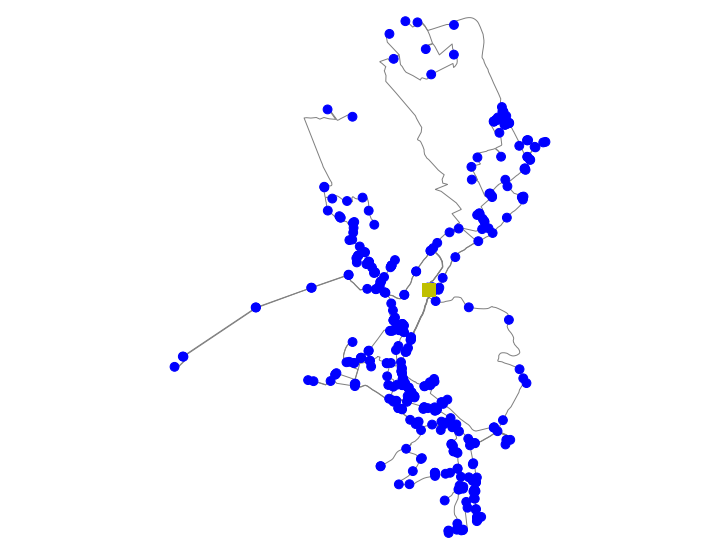}
 \caption{An illustration of the regional power grid dataset. The green block represents the electrical substation of the city and the blue circles represent the regional power consumers and generators, i.e., the prediction targets.} 
 \label{fig:grid}
\end{figure}

The regional power grid dataset~\cite{he2021application} is collected from a real-world German regional flexibility market, including two-year NWP data, the low-/medium-voltage power generation (e.g., wind and solar power generation) and consumption (e.g., residential and industrial consumption) measurements in the same period, and the geographic and electrical information of the power grid as shown in Fig.~\ref{fig:grid}.

The NWP data contains 13 24-hour-ahead numerical weather features with a 15-minute resolution.
The power data contains historical samples of 11 renewable power generators, 55 local energy consumers, and 36 low-voltage residential consumers.
Such as the NWP data, the power data ranges from March 1, 2019, to March 31, 2021, with a 15-minute resolution.
NWP and power data are scaled between 0 and 1 using min-max normalization.

The power grid information records all information regarding the regional energy market, such as the energy generators and consumers' parameters, the topological structure of the power grid, and the connection points to higher or lower power grid levels, etc.
It can help to create a virtual power grid using the open-source python library, pandapower~\cite{thurner2018pandapower}, to analyze the power grid's state and optimize power supply and demand.

Because all generators are located in the same region, the NWP features are viewed as the identical input for predicting all power targets in the power grid.
Therefore, we can build a multiple-output neural network with a shared sub-network that extracts public representations, as shown in Fig.~\ref{fig:2scenarios}, to assess the V-CLeaR framework in the DDI and TDI scenarios.
Additionally, the virtual power grid can help to analyze the effect of continually updated prediction models on power grid management and optimization.

\section{Conclusion}
\label{sec:conclusion}
Conclusively, this proposal aims to present the existing research questions related to continual deep learning for regression tasks.
Based on these questions and requirements from real-world applications, this article proposes an explainable neural-network-based CL framework for solving data-domain or target-domain incremental regression tasks.
Currently, the work is in the progress of developing the modules of this framework and evaluating the functionality of each module in the application scenario of power forecasts.
The V-CLeaR framework is expected to be modularized.
Users can utilize the single module or customize the framework for their requirements.
Our previous works have proven the applicability and necessity of the proposed framework.

\section{Acknowledgment}
This work was supervised by Prof. Dr. Bernhard Sick and supported within the Digital-Twin-Solar (03EI6024E) project, funded by BMWi: Deutsches Bundesministerium für Wirtschaft und Energie/German Federal Ministry for Economic Affairs and Energy.
\bibliographystyle{splncs04}
\bibliography{He}
\end{document}